\documentclass[letterpaper]{article}
\usepackage[utf8]{inputenc}
\usepackage{aaai22}
\usepackage{times}  
\usepackage{helvet}  
\usepackage{courier}  
\usepackage[hyphens]{url}  
\usepackage{graphicx} 
\usepackage{natbib}  
\usepackage{caption} 
\DeclareCaptionStyle{ruled}{labelfont=normalfont,labelsep=colon,strut=off} 
\usepackage{algpseudocode,algorithm,algorithmicx}
\usepackage{amsmath, amssymb}
\usepackage{arydshln}
\usepackage{subfigure}
\input{preface}
\newif\ifarxiv

\usepackage{natbib} 

\newcommand{\fairlearn}{CONFAIR }

\AtBeginDocument{%
  \providecommand\BibTeX{{%
    \normalfont B\kern-0.5em{\scshape i\kern-0.25em b}\kern-0.8em\TeX}}}

\date{February 2021}

\begin{document}
\title{
\fairlearn: Interpretable and Configurable Algorithmic Fairness}

\author{Ankit Kulshrestha \\
Ilya Safro}
\affiliations{%
  Computer and Information Science Department,
  University of Delaware,
  Newark, DE-19716
  
  \texttt{akulshr@udel.edu}
  \texttt{isafro@udel.edu}
}


\maketitle

\begin{abstract}

The rapid growth of data in the recent years has led to the development of complex learning algorithms that are often used to make decisions in real world. While the positive impact of the algorithms has been tremendous, there is a need to mitigate any bias arising from either training samples or implicit assumptions made about the data samples. This need becomes critical when algorithms are used in automated decision making systems that can hugely impact people's lives. 

Many approaches have been proposed to make  learning algorithms fair by detecting and mitigating bias in different stages of optimization. However, due to a lack of a universal definition of fairness, these algorithms optimize for a particular interpretation of fairness which makes them limited for real world use. Moreover, an underlying assumption that is common to all algorithms is the apparent equivalence of achieving fairness and removing bias. In other words, there is no user defined criteria that can be incorporated into the optimization procedure for producing a fair algorithm. Motivated by these shortcomings of existing methods, we propose the \fairlearn procedure that produces a fair algorithm by incorporating user constraints into the optimization procedure. Furthermore, we make the process interpretable by estimating the most predictive features from data. We demonstrate the efficacy of our approach on several real world datasets using different fairness criteria.\\
\noindent {\bf Reproducibility:} All source codes, and experimental data are available in our Github repository\footnote{\url{https://github.com/aicaffeinelife/CONFAIR/tree/master}}



\end{abstract}
\section{Introduction}
\label{sec:introduction}

The development of data driven expert artificial intelligence (AI) systems in the recent years has had a significant impact in the way we interact with the world around us. The ever evolving sophistication of expert systems has led to a tremendous efficiency in making predictions about diverse data and consequently alleviate tedious manual processing. If however, the predictions made by an expert system are applied directly to a person or a group of persons then serious consequences can arise. For instance, an expert system trained on biased racial data may give allocate better healthcare to a certain racial group~\citep{ziadracial} or tag more persons belonging to a certain race as relapse criminals~\citep{compasdata}. Moreover, if data exhibits gender imbalance then an expert system may faultily show targeted advertisement based on gender when it is not supposed to~\citep{datta2015automated} or allow only a certain gender to apply for a particular job~\citep{amazonsx}. 



The above real-world examples show the importance of having a fair expert system. In many situations the data collection is not in the hand of the algorithm designer and hence algorithmic fairness techniques are required to ensure a fair AI system.  The need for fair algorithms has led to the development of two parallel branches of work each with its own goal~\citep{pessach_algorithmic_2020}. One of the branches is concerned with the \emph{individual} fairness, i.e., the algorithm needs to ensure that any individual belonging to a population does not get biased decisions. The other branch has the objective of ensuring fairness amongst  \emph{sub-groups} of a given population. This objective is called statistical or group fairness. Examples of different statistical fairness definitions include Disparate Impact~\citep{chierichetti_fair_2017, ahmadian_clustering_2019, ahmadian_fair_2020, Barocas2016BigDD}, Demographic Parity~\citep{zafar_fairness_nodate, zliobaite2015relation, calders_demo_parity} and Equal Opportunity~\citep{zafar_fairness_2017, hardt_equality_2016, donini_empirical_2020}. In this paper, we exclusively focus on optimizing supervised learning algorithms in the context of statistical fairness by using support vector machine as an exemplar supervised learning.

The different definitions of statistical fairness lead to the evaluation of an algorithm's performance in fundamentally different ways. For instance, the disparate impact measures the \emph{ratio} of true positive rates across groups while equal opportunity measures the \emph{difference} between the true positive rates and false positive rates across groups with the restriction that they are less than a specified threshold. It has been shown that the different fairness metrics are often incompatible with each other~\citep{pleiss_fairness_2017, friedler2016impossibility, dwork2018group} and other complementary goals of optimization. There currently does not exist a universal definition of algorithmic fairness. This lack of a universal definition of fairness has led to the development of algorithms that optimize for only a particular definition. The resulting algorithm is only useful in scenarios where the motivating definition is applicable. Even when algorithms are optimizing for a particular fairness criteria, they make a simplifying assumption that the best way to achieve fairness is to set the fairness constraint to zero for any given data. However, these thresholds must be either inferred from the given data or supplied to the optimization process as a hyperparameter since we want the algorithm to be an \emph{equitable planner}~\citep{Kleinberg}.

\paragraph{Our contribution} The development of fair algorithms that optimize for certain definitions of fairness has hindered the development of a consistent framework for fairness. Lack of ability to parametrize and configure fairness in the existing AI expert systems represents a major gap that has to be bridged to correctly apply fairness in the real world systems. Our proposed algorithm \fairlearn takes a step in this direction, by decoupling the algorithm and the fairness criteria. We further address the lack of configurability in existing algorithms by introducing a tunable hyperparameter in our algorithm. More specifically we: 

\begin{itemize}
    \item Design a novel training procedure that is compatible with different types of classifiers. The procedure is configurable in the choice of criteria and the amount of fairness that can be injected into the model.
    
    \item Propose a data driven approach for estimating the critical predictive features in the data and utilizing the most predictive features for optimization. The discovered features also help in interpreting the predictions of the model in a much more principled manner. 
    
    \item Demonstrate the efficacy of our method on several real world datasets that are used in expert AI systems with different fairness metrics. 
    
\end{itemize}

\section{The \fairlearn Algorithm}
\label{sec:method}

We begin by describing the  notation that is used  throughout the paper. Let $\mathcal{D}$ be a dataset consisting of $n$ i.i.d. samples. Each sample in the dataset is a 3-tuple $(\textbf{x}_{i}, s_{i}, y_{i})$ where $\textbf{x}_{i} \in \mathbb{R}^{d}$ is a data point consisting of $d$ features. Associated with each data point is a label $y_{i}$ and a \emph{sensitive} feature $s_{i}$. We further assume that $y_{i} \in \{-1, 1\}$ and the sensitive feature is bi-valued i.e. $s_{i} \in \{a, b\}$. A user supplied fairness tolerance parameter $f_{tol}$ is expected to control the amount of fairness injected into a model.

 When $s_{i} \in \textbf{x}_{i}$ then most algorithms that address fairness optimize to remove the influence of $s_{i}$ over the final outcome~\citep{zafar_fairness_2017, donini_empirical_2020,dwork2012}. However, this process is flawed for two reasons. First, it assumes that all features of the data including the sensitive feature contribute equally to the final outcome. Second, it does not account for the possibility that the sensitive feature may not \emph{directly} affect the decision boundary but may indirectly influence it through another highly correlated feature. 

We hypothesize that there must exist a set of features that are critical for prediction. In other words, if any feature belonging to this set is removed from the data, the overall performance of the algorithm will decrease significantly. We call this set as the \textbf{\emph{critical feature set}}. Depending on the data, $s_{i}$ can either be present inside or outside the critical feature set. These cases give rise to different types of unfairness in a model as noted by Kamishmia~\emph{et al.}~\citep{kamishima2011fairness}. If the former case is true, then we have a case of direct bias and we cannot remove the sensitive feature from the data. However, if the latter case is true then there may still exist implicit bias caused by features having a high degree of correlation with the sensitive feature. Hence the removal of the sensitive feature does not guarantee fairness. If $s_{i} \not \in \textbf{x}_{i}$, i.e., the sensitive feature is not a part of the data point's feature then the same analysis can be applied and a critical feature set can be built either by concatenation of $s_{i}$ to $\textbf{x}_{i}$  (if $s_{i}$ is deemed to be in the critical feature set) or by building a subset of critical features that are highly correlated with $s_{i}$. \emph{One of the main features of our algorithm is that we estimate the critical feature set as the part of optimization process. The estimated critical features can be used to better interpret the underlying machine learning model.}

Algorithm~\ref{alg:fairlearn} describes our procedure in broad strokes. It consists of three main building blocks:

\begin{itemize}
    \item \texttt{FINDCRITFEATS}: This procedure determines the critical feature set from the given data and an unfair classifier.  
    \item \texttt{FINDCOVARIATES}: For a given sensitive feature $s$, this procedure estimates the covariance matrix of the data and returns a list of the features that have a high degree of correlation with $s$. 
    \item \texttt{OPTIMIZE}: Integrates the user supplied $\mathcal{F}$ into the training procedure and returns a fair classifier. 
\end{itemize}

We partition $\mathcal{D}$ into a training set $\mathcal{D}_{\text{train}}$ and a test set  $\mathcal{D}_{\text{test}}$. An unfair classifier $\mathbb{A}$ is used to estimate the critical feature set. We supply as inputs to our procedure $\mathcal{D}_{\text{train}}, \mathcal{D}_{\text{test}}$, and  $\mathbb{A}$ along with a fairness criteria $\mathcal{F}$,the index of sensitive feature such that $\textbf{x}[i] = s_{i}$ and a fairness threshold $f_{tol}$. 

Our procedure (\fairlearn) begins by estimating the critical feature set $\mathcal{C}$ and a set of correlated features $\psi$ from the given data (Lines~\ref{lin:fcrit} and Lines~\ref{lin:fcov}). The set $\mathcal{C}$ is sorted in descending order based on the magnitude of impact a given feature has on the accuracy. The set $\psi$ is sorted in descending order based on the magnitude of covariance with the protected feature. If $s_{i}$ is within $\mathcal{C}$, then we run the \texttt{OPTIMIZE} procedure  with the objective of producing a fairer classifier $\mathcal{A}$ from $\mathbb{A}$ (Line~\ref{lin:optimize_c1}). If $s_{i}$ is not found with $\mathcal{C}$, then we cannot assume the model is fair. This is because, the sensitive feature can have a highly corelated feature which may be a part of the critical feature set. Hence, we check if any of its covariates are present in $\mathcal{C}$. We build a new critical feature set $\mathcal{C}'$ and then run \texttt{OPTIMIZE} (Line~\ref{lin:optimize_c2}). If however, neither case is true then we can conclude the source of  bias is \emph{extrinsic} to the model and fairness can be achieved with data preprocessing techniques. We describe the different subroutines in the following subsections.  

\begin{algorithm}
\caption{The \fairlearn Algorithm}
\label{alg:fairlearn}
\begin{algorithmic}[1]
\Procedure{\fairlearn}{$\mathcal{D}_{\text{train}}, \mathcal{D}_{\text{test}}, \mathbb{A}, s, \mathcal{F},
f_{tol}$}
\State $\mathcal{C} \gets \Call{FINDCRITFEATS}{\mathcal{D}_{\text{train}}, \mathcal{D}_{\text{test}}, \mathbb{A}}$\label{lin:fcrit}
\State $\psi \gets \Call{FINDCOVARIATES}{\mathcal{D}_{\text{train}}, s}$\label{lin:fcov}
\If{$s \in \mathcal{C}$}
    \State $\mathcal{A} \gets \Call{OPTIMIZE}{\mathcal{D}_{\text{train}}, \mathcal{C}, \mathcal{F}, f_{tol}}$ \label{lin:optimize_c1}
\ElsIf{$\psi \neq \emptyset$ \& $\texttt{Any}(\psi) \in \mathcal{C}$}
    \State $\mathcal{C}' \gets \psi \cap \mathcal{C}$ \label{lin:ncritset} \Comment{intersection of covariates and critical features.}
    \State $\mathcal{A} \gets \Call{OPTIMIZE}{\mathcal{D}_{\text{train}}, \mathcal{C}', 
    \mathcal{F}, f_{tol}}$ \label{lin:optimize_c2}
\Else
    \State $\mathcal{A} \gets \mathbb{A}$
\EndIf
\State \textbf{return} $\mathcal{A}$
\EndProcedure
\end{algorithmic}

\end{algorithm}

\subsection{Finding Critical Features and Covariates}
The first step in \fairlearn is to determine the critical features and the sensitive features covariates. We show our approach for this step in Algorithms~\ref{alg:crit_feats_covs} and Algorithm~\ref{alg:find_covariates}.

\noindent
The main idea in the \texttt{FINDCRITFEATS} procedure is to leverage \emph{permutation feature importance}~\citep{Breiman2001} for estimating the overall importance of a particular feature. To effectively test the importance of a feature, we use the accuracy on the test set as a measure of performance for the given classifier. However, we stress that nothing prevents using other performance measures that might be more suitable for specific tasks and data. We accept a parameter $K$ that determines the number of permutations to perform per feature. This parameter must be chosen carefully since too many permutations can impact the running time of the algorithm, especially for large datasets. For each feature $j$ we permute it $K$ times and measure the score on the test set (Line~\ref{lin:shuffled_acc}). A permutation of feature $j$ is defined as a shuffling of data points that breaks the association between feature $j$ and the observed outcome (Line~\ref{lin:rand_shuffle}). The importance of a feature $j$ is then given by $\Delta - \frac{1}{K} \sum_{i=1}^{K} \delta_{k}$, where $\Delta$ is the accuracy of the model when no features were permuted, $\delta_{k}$ is the accuracy of the model at the $k^{th}$ permutation step. At the end of the procedure we sort  the indices in the descending order of importance and return the indices to the main algorithm (Line~\ref{lin:fcrit_sort}).

In order to find the set of features that are correlated with $s$ from data, we adopt a Maximum Likelihood approach that aims to fit an estimator to determine the covariance matrix. We assume that all examples are i.i.d and the number of samples is much larger than the number of features in the data. Once the covariance matrix $\varepsilon$ is determined (Line~\ref{lin:mle_estimator}) we compute the sensitive feature covariates $\varepsilon_{s}$. We reject the features that are negatively correlated. We further reject the sensitive feature and return the indices of the remaining covariates sorted in descending order (Line~\ref{lin:fcov_sort}).
\begin{algorithm}
\caption{The FINDCRITFEATS Procedure}\label{alg:crit_feats_covs}
\begin{algorithmic}[1]
\Procedure{FINDCRITFEATS}{$\mathcal{D}_{\text{train}}, \mathcal{D}_{\text{test}}, \mathbb{A}, K$}
    \State $\mathbb{A} \gets \Call{TRAIN}{(X,Y) \in \mathcal{D}_{\text{train}}}$
    \State $\mathcal{N} \gets \Call{DIM}{\mathcal{D}_{\text{train}}}$ 
    \State $(X_{\text{test}},
        Y_{\text{test}}) \gets \mathcal{D}_{\text{test}}$
    \State $I \gets \{0, 0, \dots, 0\}$ \Comment{Feature importance list for $\mathcal{N}$ features}
    \State $\Delta \gets \Call{ACCURACY}{\mathbb{A}, X_{\text{test}}, Y_{\text{test}}}$ \label{lin:base_acc_score}
    \For{$j \in \mathcal{N}$}
      \State Scores $\gets \emptyset$
      \For{$k \in RANGE(K)$}
        \State $\hat{X}_{\text{test}} \gets \Call{RANDOMSHUFFLE}{X_{\text{test}}, j}$ \label{lin:rand_shuffle}
        \State $\delta_{k} \gets 
        \Call{ACCURACY}{\mathbb{A}, \hat{X}_{\text{test}}, Y_{\text{test}}}$ \label{lin:shuffled_acc}
        \State Scores $ \gets \text{Scores} \cup \delta_{k}$
      \EndFor
      
      \State $I_{j} \gets  \Call{IMPORTANCE}{\Delta, \text{Scores}}$ \label{lin:imp_sampling}
    \EndFor
    
    \State \textbf{return} $\Call{ARGSORT}{I}$ \label{lin:fcrit_sort}
\EndProcedure

\end{algorithmic}

\end{algorithm}
\begin{algorithm}
\caption{The FINDCOVARIATES Procedure}\label{alg:find_covariates}
\begin{algorithmic}[1]
\Procedure{FINDCOVARIATES}{$\mathcal{D}_{\text{train}}, s$}
    \State $\varepsilon \gets \Call{MLE\_ESTIMATOR}{\mathcal{D}_{\text{train}}}$ \label{lin:mle_estimator}
    \State $\varepsilon_{s} \gets \varepsilon[:, s]$
    \State $C \gets \varepsilon_{s} > 0.0$
    \State \textbf{return} $\Call{ARGSORT}{C}$[0:] \label{lin:fcov_sort} \Comment{Return all covariate features except the first.}
\EndProcedure
\end{algorithmic}
\end{algorithm}



\subsection{The OPTIMIZE Procedure}

The \texttt{OPTIMIZE} accepts the training set $\mathcal{D}_{\text{train}}$, the critical feature set $\mathcal{C}$, a fairness criteria $\mathcal{F}$ and a tolerance parameter $f_{tol}$. The main goal of this procedure is to provide a consistent framework for finding a fairer algorithm $\mathcal{A}$ by optimizing for any $\mathcal{F}$. It is a general procedure in the sense that,  a user  may supply their own \texttt{OPTIMIZE} procedure depending upon the choice of unfair algorithm $\mathbb{A}$. The only restriction is that the supplied procedure must be  parametrized in a similar manner to the one shown in Algorithm~\ref{alg:fairlearn}.  In such a case our algorithm will still find the best features for the data and return a fairer algorithm according to the provided optimization procedure. For instance, if one wishes to optimize a multi-layer perceptron on such fairness metric as the Equal Opportunity, she can provide an \texttt{OPTIMIZE} procedure that integrates the fairness constraint as a regularization term in the loss function.  

\subsubsection{Optimizing with Support Vector Machines}

We develop an instance of \texttt{OPTIMIZE} procedure for a support vector machine (SVM) with the Equal Opportunity (EO) fairness metric. We largely adopt the framework of Donini \emph{et al.}~\citep{donini_empirical_2020} in the design of the optimization procedure with fairness constraints with some important modifications in the formulation. We translate the  definition of the EO metric, i.e.,
\begin{equation}
\abs[\Big]{P[D=1 | \mathcal{S}=a, \mathcal{Y}=1] - P[D=1 | \mathcal{S}=b, \mathcal{Y}=1]} \leq \epsilon,
\end{equation}
where $D$ is the predicted outcome, into a convex constraint:

\begin{equation}
  \abs[\Big]{\mathbb{E}\left[l^{a}(f(\textbf{x}), y_{+}) - l^{b}(f(\textbf{x}, y_{+})\right]} \leq \epsilon.
    \label{eq:ferm_constraint}
\end{equation}

The convex and smooth loss function $l^{g}(f, y_{+})$ in Equation~\ref{eq:ferm_constraint} measures the performance of a family of functions $f$ in predicting positively labeled samples $y_+$ belonging to a group $g = \{a,b\}$. The overall goal of optimization is to find 
\begin{equation}
    \text{min}\left\{f \in \mathcal{H} ; \textrm{s.t.}\   \mathbb{E}[l^{a}(f, y_{+}) - l^{b}(f, y_{+})]] \leq \epsilon \right\},
\end{equation}
where $\mathcal{H}$ is the hypothesis space of functions.

As a representative example, we choose $f(\textbf{x}) = \langle \textbf{w}, \phi(\textbf{x}) \rangle$ and assume that the underlying hypothesis space $\mathcal{H}$ is a Reproducing Kernel Hilbert Space (RKHS). This assumption allows us to define a kernel mapping in high dimension space $K(\textbf{X}, \textbf{X}') = \langle \phi(\textbf{X}), \phi(\textbf{X}') \rangle$. In~\citep{donini_empirical_2020}, a parameter $\textbf{u}_{g}$ is introduced that measures the barycenter of the induced feature mapping as:

\begin{equation}
    \textbf{u}_{g} = \frac{1}{N^{+}_{g}} \sum_{j \in I^{+}_{g}} \phi(x_{j})
    \label{eq:base_bary}
\end{equation}
In Equation~\ref{eq:base_bary}, $N^{+}_{g}$ is the number of positively labeled points belonging to the sensitive group $g$ and $I^{+}_{g}$ is the set of indices of such points. Based on the definition of $\textbf{u}_{g}$ in Equation~\ref{eq:base_bary}, the fairness constraint was specified as $\langle \textbf{w}, |\textbf{u}_{a} - \textbf{u}_{b}| \rangle \leq \epsilon$. The constraint can be understood as a way to minimize the effect of the sensitive group clusters on the overall decision boundary. This formulation is effective in injecting fairness in model, but only matches the first order moments of the clusters. We improve upon this approach by taking into account the \emph{spread} of the respective clusters. Specifically, we redefine $\textbf{u}_{g}$ as:


\begin{equation}
    \hat{\textbf{u}}_{g} = \frac{1}{\sigma_{g}} \left[\frac{1}{N^{+}_{g}} \sum_{j \in I^{+}_{g}}\phi(x_{j})\right]
    \label{eq:our_contri}
\end{equation}

\noindent
In Equation~\ref{eq:our_contri}, $\sigma_{\hat{g}}$ is the square root of inter-cluster variance:

\begin{equation}
    \sigma_{g}^{2} = \frac{1}{N^{+}_{g} - 1} \sum_{i=1}^{n} \left[\phi(x_{i})^{+}_{g} - \mu_{\hat{g}}\right]^{2}
    \label{eq:scaling-fac}
\end{equation}

\noindent
In Equation~\ref{eq:scaling-fac}, $\mu_{\hat{g}}$ is the mean of the positively labeled points in a group different from the current one. In our experiments we found that this scaling significantly improved the fairness of our model. Based on this modification, the constrained optimization objective is stated as follows:

\begin{equation}
    \begin{aligned}
       &\min_{\gamma \in \mathbb{R}^{n}}  \sum_{i=1}^{n} l\left(\sum_{j=1}^{n} K_{ij}\gamma_{i}, y_{i}\right) + \lambda \sum_{(i,j)=1}^{n} \gamma_{i}\gamma_{j}K_{ij} \\ 
       &\textrm{s.t.}  \left|\sum_{i=1}^{n} \gamma_{i}[\frac{1}{\sigma_{a}}(\frac{1}{N^{+}_{a}}\sum_{j \in I^{+}_{a}} K_{ij}) - 
       \frac{1}{\sigma_{b}}(\frac{1}{N^{+}_{b}}\sum_{j \in I^{+}_{b}} K_{ij})]\right| \leq f_{tol}
    \end{aligned}
    \label{eq:svm_objective}
\end{equation}
In Equation~\ref{eq:svm_objective}, we make use of the Representer theorem~\citep{kimeldorf1971some} to express $\textbf{w} = \sum_{i=1}^{n} \gamma_{i} \phi(x_{i})$, where $\gamma \in \mathbb{R}^{n}$ and solve the Tikhonov regularization problem. 

The entire setup effectively translates to solving the dual of a soft-margin SVM with the fairness constraint. \emph{It is important to mention that in contrast with the approach in~\citep{donini_empirical_2020}, we do not restrict ourselves to the case where $f_{tol} = 0$ and allow it to be a user defined parameter which is extremely important for the real-world applications.} 

Algorithm~\ref{alg:minimize} shows the procedure for solving the dual of the optimization problem in Equation~\ref{eq:svm_objective}. The variables $\vartheta_{a}, \vartheta_{b}$ are selected based on the fairness criteria and can be a set of either positively or negatively labeled points. The matrices $\textbf{G}, \textbf{h}$ encode the box constraint $0 \leq \vec{\alpha} \leq C$. For simplicity, we add the fairness criteria to the affine constraint in a quadratic program. We make use of an off-the-shelf quadratic solver (e.g.,  CVXOPT or MOSEK~\citep{mosek}). The resulting vector of solutions $\vec{\alpha}$ is used to get the support vectors and make predictions via the \texttt{GET\_ALG} function.

\begin{algorithm}
\caption{OPTIMIZE with SVM Objective }\label{alg:minimize}
\begin{algorithmic}[1]
 \Procedure{OPTIMIZE}{$\mathcal{D}_{\text{train}}, \mathcal{C}, \mathcal{F}, f_{tol}, s$}
  
  \State $s \gets \Call{FINDINDEX}{\mathcal{C}, s}$
  \State $\textbf{X}, \textbf{y} \gets \mathcal{D}_{\text{train}}$
  \State $\textbf{X} \gets \textbf{X}[:, \mathcal{C}]$
  \State $N \gets |\mathcal{D}_{\text{train}}|$
  \State $\vartheta_{a} \gets \mathcal{I}^{\mathcal{F}}_{a}$ \Comment{Set of all indexes with sign $\mathcal{F}$ and group $a$}
  \State $\vartheta_{b} \gets \mathcal{I}^{\mathcal{F}}_{b}$ \Comment{Set of all indexes with sign $\mathcal{F}$ and group $b$}
  
  \State $K \gets \Call{KERNEL}{\textbf{X}, \textbf{X}}$ 
  \State $\textbf{P} \gets \Call{OUTER}{y,y} * K$ \Comment{Setting up optimization problem}
  \State $\textbf{q} \gets -1 * \textbf{y}$
  \State $\textbf{G} \gets \left[\mathbb{D}^{N}, \textbf{I} \right]$
  \State $\textbf{h} \gets [0, -C]$ \Comment{$C$ is the  control parameter in soft SVM}
  \State $\textbf{A} \gets \left[\textbf{y}, \textbf{y}*|\hat{\textbf{u}}_{a} - \hat{\textbf{u}_{b}}|\right]$ \Comment{Fairness constraint}
  \State $\textbf{b} \gets [0, f_{\text{tol}}]$
  \State $\vec{\alpha} \gets \Call{CONEQP}{\textbf{P}, \textbf{q}, \textbf{G}, \textbf{h}, \textbf{A},
  \textbf{b}} $
  
  \State \textbf{return} $\Call{GET\_ALG}{\vec{\alpha}}$
\EndProcedure
\end{algorithmic}
\end{algorithm}

\section{Experiments}
\label{sec:experiments}

Implementation of \fairlearn and data are available in \citep{fairlearn:repo}. The goal of our experiments is threefold. First, we want to establish that \fairlearn  achieves fairness given any fairness criteria without suffering a significant degradation in accuracy. Second, we wish to study the effect of the user supplied $f_{tol}$ values on different data. Third, we want to demonstrate the generality of our approach by showing that  \fairlearn works in a model and data agnostic manner. For all our experiments we use the \texttt{OPTIMIZE} procedure developed for SVM in the previous section.

\begin{table}[ht]
    \centering
    \begin{tabular}{|l|c|c|c|}
    \hline
        Dataset  & \# Data & \# Features & Sensitive  \\
                  & points &  & Attribute  \\
    \hline
        Adult & $32561$ & $12$ & Gender  \\
        COMPAS & $6171$ & $10$ & Race \\ 
        German & $1000$ & $59$ & Gender \\
    \hline
    \end{tabular}
    \caption{Statistics of real world datasets used in experiments.}
    \label{tab:dataset_stats}
\end{table}



    
    

A description of these datasets is provided in Table~\ref{tab:dataset_stats}. For all datasets,we perform a 10-fold cross validation to determine the optimal hyperparamters $C$ and $\gamma$ for non-linear SVM. For all datasets except the Adult dataset, we repeat this procedure over five different runs and report the average accuracy with their standard deviation. We use the provided train/test split for the Adult dataset and report the accuracy over the test set. \emph{Unlike previous work in this area, we evaluate our method on different values of $f_{tol}$.} To model a realistic scenario, one critical assumption we make in all experiments is that the sensitive parameter $s$ is always inside the training data.

\subsection{Fairness Metrics}
For a supervised learning paradigm, we define three distinct variables. $D$ is the predicted variable by a supervised learning algorithm, $\mathcal{Y}$ is the ground truth label and $S$ is the sensitive feature of the data point. Fairness criteria for the supervised paradigm can be divided into two broad categories depending upon the choice of the conditioning variable. 

The first category is a set of metrics that are conditional on the observed variable $\mathcal{Y}$. This is equivalent to saying $D \perp \mathcal{S} | \mathcal{Y}$, where $D \perp \mathcal{S}$ implies that $D$ is independent of $\mathcal{S}$. Some criteria that fall in this category are Equal Odds (where $D = -1 \perp \mathcal{S} | \mathcal{Y} = 1$ and $D = 1 \perp \mathcal{S} | \mathcal{Y}=1$)~\citep{hardt_equality_2016}, and Equal Opportunity (where $D = 1 \perp \mathcal{S} | Y=1$)~\citep{donini_empirical_2020, zafar_fairness_2017}. A defining characteristic of these metrics is that they  can be interpreted as a score function $\mathcal{L}: \mathcal{X} \times \mathcal{S} \times \mathcal{Y} \rightarrow \mathbb{R}$. For instance, the Equal Opportunity metric can be expressed as
\begin{equation}
  \mathcal{L}(\mathcal{X}, \mathcal{Y},  \mathcal{S}) =  \mathbb{E}\left[|l^{+}_{a}(f(\textbf{x}_{i}), y_{i}) - l^{+}_{b}(f(\textbf{x}_{i}), y_{i})|\right]
  \label{eq:deo_loss}
\end{equation}
and a corresponding score function for Equality of True Negative Rates can be written as
\begin{equation}
  \mathcal{L}(\mathcal{X}, \mathcal{Y},  \mathcal{S}) =  \mathbb{E}\left[|l^{-}_{a}(f(\textbf{x}_{i}), y_{i}) - l^{-}_{b}(f(\textbf{x}_{i}), y_{i})|\right]
  \label{eq:dtnr_loss}
\end{equation}

\noindent
In Equation~\ref{eq:deo_loss} and~\ref{eq:dtnr_loss}, $l^{\pm}_{g}$ is the per sample loss for any learning algorithm $f$ on either positively or negatively labeled points belonging to a sensitive group $g$.
In this paper we consider the Equal Opportunity as a representative example of fairness from this category. Specifically, we evaluate the Difference of Equal Opportunity(DEO) metric~\citep{donini_empirical_2020} 
\begin{equation}
\abs[\Big]{P\left[D = 1 | \mathcal{S}=a,\mathcal{Y}=1\right] - P\left[D = 1 | \mathcal{S}=b, \mathcal{Y}=1 \right]} \leq \epsilon.
\label{eq:deo_metric}
\end{equation}

\noindent
The second category is a set of metrics that are conditional on the \emph{predicted} outcome $D$. This is equivalent to saying $\mathcal{Y} \perp \mathcal{S} | D$. Some examples of metrics belonging to this class are Equality of Negative Predictive Value $Y = -1 \perp \mathcal{S} | D=-1$ and Equality of Positive Predictive Value $\mathcal{Y} = 1 \perp \mathcal{S} | D=1$. For a more thorough treatment of these metrics and the ones discussed above we refer an interested reader to~\citep{Mitchell_2021}. These metrics quantify the \emph{predictive quality} of a trained algorithm. This is important since a deployed system rarely has access to true outcomes. If an algorithm has a good predictive quality then we can be confident that the decisions made by the algorithm more or less reflect the actual observed outcomes. In this paper we choose the Equality of Negative Predictive Value (NPV) defined as 
\begin{equation}
\abs[\Big]{P\left[\mathcal{Y} = -1 | \mathcal{S}=a, D=-1\right] - P\left[\mathcal{Y} = -1 | \mathcal{S}=b, D=-1 \right]} \leq \epsilon.
\label{eq:npv_metric}
\end{equation}

For both metrics in Equation~\ref{eq:deo_metric} and Equation~\ref{eq:npv_metric} a lower value signifies a higher amount of fairness in a model.

\begin{figure}[t]
    \centering
    \includegraphics[width=.75\columnwidth]{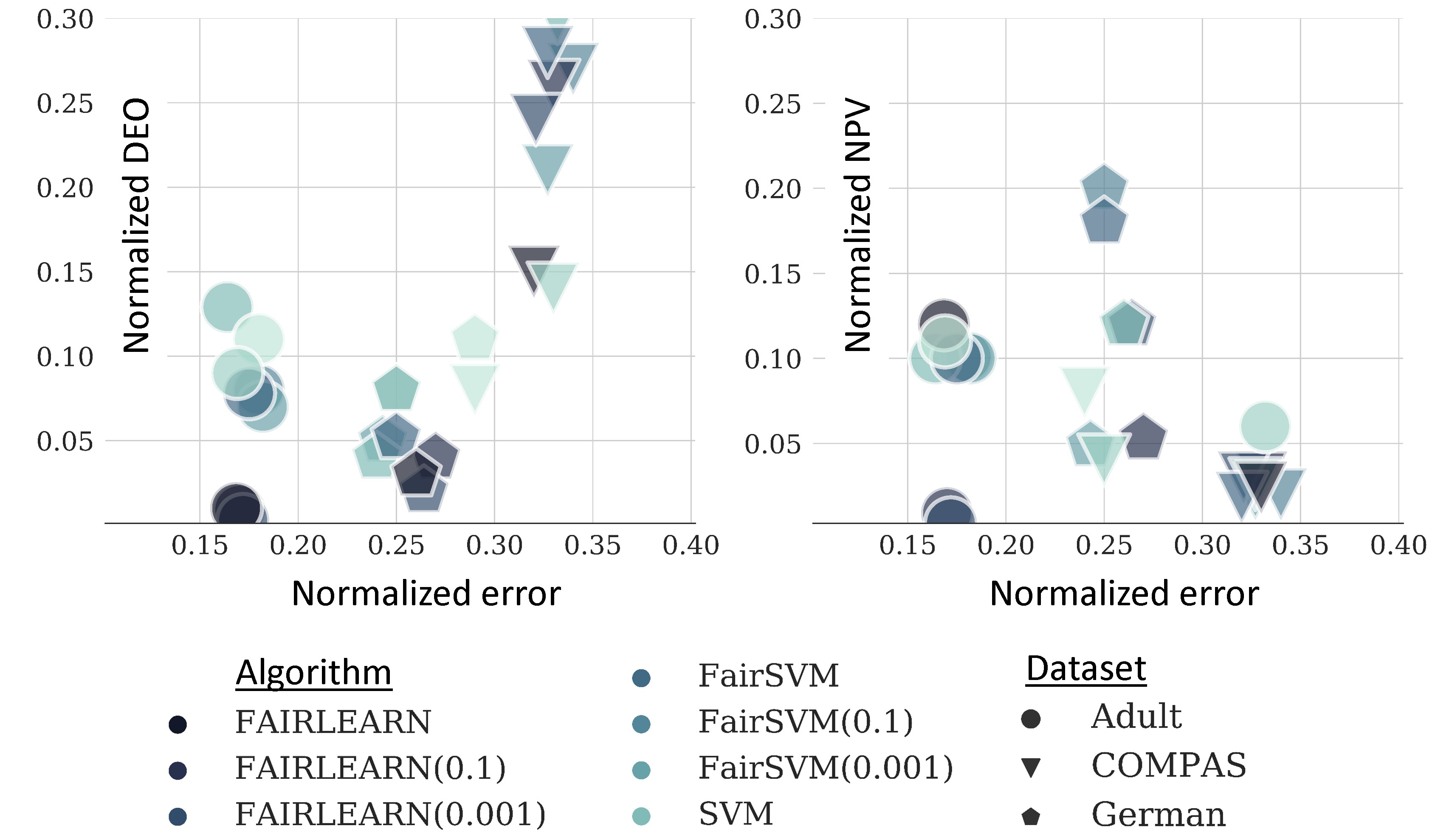}
    
    \caption{Comparison of \fairlearn classifier with other methods on different metrics. Left: DEO, Right: NPV}
    \label{fig:alg_compare}
\end{figure}

\subsection{Experimental Evaluation}

We now present our results on the three real world datasets discussed earlier. Based on the constrained optimization objective in Equation~\ref{eq:svm_objective}, we divide our experiments across two categories - a fair SVM trained with our dual formulation on the full feature set and a fair SVM trained with our proposed \fairlearn procedure. We call the former SVM as \emph{FairSVM} and the latter as \emph{\fairlearn}. We measure the performance of both algorithms with three baseline methods - an unfair SVM trained to optimize accuracy, the FERM method by Donini~\emph{et al.}~\citep{donini_empirical_2020} and the post processing step proposed by Hardt~\emph{et al.}~\citep{hardt_equality_2016}. For the baseline results of Hardt~\emph{et al.} we report the results obtained in~\citep{donini_empirical_2020}. We reproduce the results for the FERM method baseline by using the code provided by the authors. We report our results for both linear and non-linear cases. For the non-linear case we use the RBF kernel  $K(\textbf{X}, \textbf{X}') = e^{-\gamma ||\textbf{X} - \textbf{X}'||^{2}}$.\\

\noindent
Table~\ref{tab:fairlearn_baseline} shows the results of our experiments. The results show a consistent improvement in the DEO objective for both FairSVM and \fairlearn classifiers across all datasets. Additionally, we note that the \fairlearn classifier with the non-linear kernel does not suffer a significant decrease in accuracy and improves upon the DEO metric, achieving the best performance amongst all methods on the German and Adult datasets in the non-linear case. This result suggests that for critical features discovered by our algorithm are useful and using them does not significantly lower the classification performance of the algorithm and leads to a much more fairer version of the same. 
In the case of linear kernel as well, our method performs much better than baselines on both Adult and German datasets in terms of fairness performance. We observe a significant drop in accuracy on the GERMAN dataset for the linear case and we feature this drop to the linear kernel not being the best choice for critical features selected by the algorithm. On the Compas dataset, our results are close to the best performing FERM method on the DEO metric. For the NPV metric, the results suggest that a better performance may be reached if we use the FairSVM method since it consistently achieves a lower score on the different datasets.

The results also show an interesting relationship between the two metrics. We note that for datasets where an unfair algorithm had a low DEO, the fairer version of the algorithms had a much higher NPV rate. However, when the unfair algorithm had a higher DEO, the our proposed fair algorithms achieved a much lower NPV rate along with the DEO \emph{without being explicitly optimized for the former metric}. These observations suggest that there may be an implicit relation between the metrics that are conditioned on the observed variable and the ones that are conditioned on the predicted variable. In a broader sense, our results showcase the trade-offs that need to be made depending on the real world situation.\\

\noindent
It is helpful to understand the relationship between the error rate of an algorithm and the fairness metrics since it helps to establish a quantitative understanding of achievable fairness at a desired error rate. Figure~\ref{fig:alg_compare} shows the normalized value of fairness metrics along with the error rate at which this value was achieved. For the Adult dataset, we observe that we can get a fair classifier with a minimal loss in accuracy. For the German credit dataset and the COMPAS dataset the decision to use a classifier is more nuanced since different classifiers (including ours with different $f_{tol}$) can attain different fairness values at different error  rates. For instance, if DEO is the fairness metric of choice, then for German dataset our method provides a lower value with a slightly higher error rate than by the classifier trained according to~\citep{donini_empirical_2020}.To the best of our knowledge, this is the first time any work has compared the effect of two different metrics at different fairness thresholds across different datasets. 


\begin{table*}[t]
    \centering
    \tiny
    \setlength{\tabcolsep}{.5pt}
    \begin{tabular}{@{}l|ccc|ccc|ccc@{}}
    \hline\hline 
        Method & \multicolumn{3}{c|}{GERMAN} & \multicolumn{3}{c|}{COMPAS} &
        \multicolumn{3}{c}{ADULT} \\ 
        \hline
        & Acc. & DEO & NPV &  Acc. & DEO & NPV &  Acc. & DEO & NPV \\
    \hline
    SVM  &  0.76 $\pm$ 0.06 & 0.04 $\pm$ 0.02 & 0.08 $\pm$ 0.04 &  0.67 $\pm$ 0.02 & 0.30 $\pm$ 0.08 & 0.06 $\pm$ 0.03 & $0.83$ & $0.129$ & $0.10$  \\
    
    FairSVM  & 0.75 $\pm$ 0.01 & 0.05 $\pm$ 0.02 &\textbf{0.04} $\pm$ \textbf{0.04} &  $0.67 \pm 0.02$ & $0.21 \pm 0.06$ & $0.026 \pm 0.01$ &  $0.824$ & $0.071$ & \textbf{0.10}  \\
     
    \fairlearn & $0.736 \pm 0.04 $ & \textbf{0.024} $\pm$ \textbf{0.01} & $0.12 \pm 0.07$ &  $0.68 \pm 0.006$ & $0.15 \pm 0.023$ & \textbf{0.018} $\pm$ \textbf{0.01} &  $0.831$ & \textbf{0.003} & $0.127$ \\
    
    FERM & $0.75 \pm 0.014$ & $0.083 \pm 0.045$ &$0.18 \pm 0.11$ & $0.67 \pm 0.01$ & \textbf{0.14}$ \pm$ \textbf{0.026} & $0.024 \pm 0.021$& 0.831 & 0.09 & 0.11\\
    
    Hardt & $0.71 \pm 0.03$ & $0.11 \pm 0.18$ & - & $0.71 \pm 0.01$ & $0.08 \pm 0.01$ & - & 0.82 & 0.11 &- \\
    \hline\hline
    
    Lin. SVM  &  $0.767 \pm 0.02$ & $0.08 \pm 0.06$ & $0.11 \pm 0.05$ &  $0.65 \pm 0.01$ & $0.244 \pm 0.57$ & $0.03 \pm 0.02$ & $0.80$ & $0.013$ & $0.17$  \\
    
    Lin. FairSVM  &  $0.74 \pm 0.02$ & \textbf{0.04} $\pm$ \textbf{0.02} & $0.12 \pm 0.06$ &  $0.65 \pm 0.01$ & $0.15 \pm 0.03$ & $0.03 \pm 0.02$ & $0.803$ & \textbf{0.003} & \textbf{0.16}  \\
     
     Lin. \fairlearn &  $0.701 \pm 0.03$ & $0.05 \pm 0.1$ & \textbf{0.10} $\pm$ \textbf{0.09} &  $0.65 \pm 0.01$ & $0.153 \pm 0.05 $ & $0.02 \pm 0.01$ &  0.78 &  0.03 & 0.17 \\
     
     Lin. FERM & $0.74 \pm 0.02$ & $0.072 \pm 0.04$ & $0.17 \pm 0.063$ & $0.65 \pm 0.01$ & \textbf{0.09} $\pm$ \textbf{0.038} & $0.038 \pm 0.02$ & 0.80 & 0.01 & 0.165 \\
     
     Lin. Hardt

     & $0.61 \pm 0.15$ & $0.15 \pm 0.13$ & - & $0.67 \pm 0.03$ & $0.21 \pm 0.09$ & - & 0.80 & 0.10 &- \\
     \hline
      
    \end{tabular}
    
    \caption{Results for all datasets on the accuracy, DEO and NPV metrics. In dataset where an explicit test set is not provided, the results are reported as mean $\pm$ standard deviation of five independent runs. Best results are shown in bold.}
    \label{tab:fairlearn_baseline}
\end{table*}
\begin{table*}[t]
    
    \centering
    \tiny
    \begin{tabular}{c | c | ccc | ccc }
        \hline\hline
        DATASET & $f_{tol}$ & \multicolumn{3}{c|}{Fair SVM} & \multicolumn{3}{c}{\fairlearn}  \\
        \hline
                & & Acc & DEO & NPV & Acc & DEO & NPV  \\ 
        \hline 
        
          & 0 &  0.82 & 0.07 & 0.101 & 0.83 & 0.003 & 0.127 \\
        ADULT & 0.1 & 0.82 & 0.08 & 0.101 & 0.83 & 0.009 & 0.126  \\
              & 0.001 & 0.82 & 0.07 & 0.10 & 0.83 & 0.003 & 0.127 \\
        \hdashline
         Baseline & - &  0.835 & 0.129 & 0.1117 & 0.835 & 0.129 & 0.117\\
        \hline
        
         & 0 & 0.67 $\pm$ 0.01 & 0.28 $\pm$ 0.04 & 0.023 $\pm$ 0.01 & 0.68 $\pm$ 0.01 & 0.15 $\pm$ 0.02 & 0.023 $\pm$ 0.01  \\ 
        COMPAS & 0.1 & 0.66 $\pm$ 0.01  & 0.27 $\pm$ 0.05 & 0.02 $\pm$ 0.01 & 0.67 $\pm$ 0.01 &   0.26 $\pm$ 0.09 & 0.03 $\pm$ 0.02 \\ 
          & 0.001 & 0.673 $\pm$ 0.01 & 0.21 $\pm$ 0.07 & 0.02 $\pm$ 0.01 & 0.679 $\pm$ 0.01 & 0.24 $\pm$ 0.07 & 0.03 $\pm$ 0.02 \\ 
        \hdashline
         Baseline & - &  0.668 $\pm$ 0.02 & 0.30 $\pm$ 0.08 & 0.06 $\pm$ 0.03 & 0.668 $\pm$ 0.02 & 0.30 $\pm$ 0.08 & 0.06 $\pm$ 0.03 \\
        \hline
        
        & 0 & 0.76 $\pm$ 0.03 & 0.05 $\pm$ 0.04 & 0.12 $\pm$ 0.06 & 0.73 $\pm$ 0.02 & 0.03 $\pm$ 0.02 & 0.27 $\pm$ 0.15  \\ 
       GERMAN  & 0.1 & 0.75 $\pm$ 0.02 & 0.08 $\pm$ 0.07 & 0.20 $\pm$ 0.078 & 0.73 $\pm$ 0.04 & 0.04 $\pm$ 0.072 & 0.053 $\pm$ 0.05  \\
       & 0.001 & 0.757 $\pm$ 0.01 & 0.05 $\pm$ 0.02 & 0.049 $\pm$ 0.04 & 0.736 $\pm$ 0.04 & 0.02 $\pm$ 0.02 & 0.12 $\pm$ 0.07\\
       \hdashline
         Baseline & - &  0.767 $\pm$ 0.068 & 0.04 $\pm$ 0.02 & 0.08 $\pm$ 0.04 & 0.767 $\pm$ 0.068 & 0.04 $\pm$ 0.02 & 0.08 $\pm$ 0.04\\
       \hline
    \end{tabular}
    
    
    \caption{Effect of $f_{tol}$ on fairness on algorithm trained with full feature and with critical set for the non-linear case. Baseline is a biased SVM trained on the same dataset.}
    \label{tab:ftol_results}
\end{table*}


\subsection{Effect of Tolerance Parameter}

We introduced the fairness tolerance parameter $f_{{tol}}$ as a way to allow a user to control the amount of fairness in a model. For the SVM dual optimization objective, this parameter translates to the allowed ``influence" of the sensitive groups over the final decisions. Our goals in this experiment were to see if the change in $f_{tol}$ changes the fairness of the model and to quantify the effect of retaining only the critical feature set against the baseline of an unfair SVM trained with exactly the same parameter grid. We compare the two classifiers that we discussed earlier. For both classifiers we used the RBF kernel and performed a 10-fold cross validation procedure. We averaged the results over five independent runs on the COMPAS and the German Dataset.\\
\noindent
\vspace{-5pt}
    
Table~\ref{tab:ftol_results} shows the results of our experiments. We can see that at all thresholds both algorithms significantly improve in the DEO metric for both Adult and Compas datasets. The case of the German dataset is interesting since the DEO metric does not uniformly decrease from the baseline for the classifier trained with full features. One reason for this observation can be the low DEO of the baseline algorithm on the German dataset. In the other two, this metric is higher in the baseline which implies a higher percentage of misclassified samples across sensitive groups. The classifier trained with our proposed procedure either matches or reduces the DEO. We also evaluate the classifiers on the NPV metric. The classifier trained with full features shows a consistent reduction from baseline across the thresholds. However, classifier trained with the \fairlearn procedure shows a consistent decrease only in the case of German and Compas datasets. For the Adult dataset, the NPV increases across thresholds. We hypothesize that the results are a combination of two factors. First, the NPV and DEO appear to be inversely related and the DEO decreases sharply. Second, if the sensitive parameter happens to be a part of the critical feature set, then the algorithm may aggressively optimize for the provided fairness constraint. We shall examine the second hypothesis in some detail in the next subsection.

\subsection{Critical Features of Different Datasets}
\begin{figure}[h]
    \centering
     \includegraphics[width=.5\columnwidth]{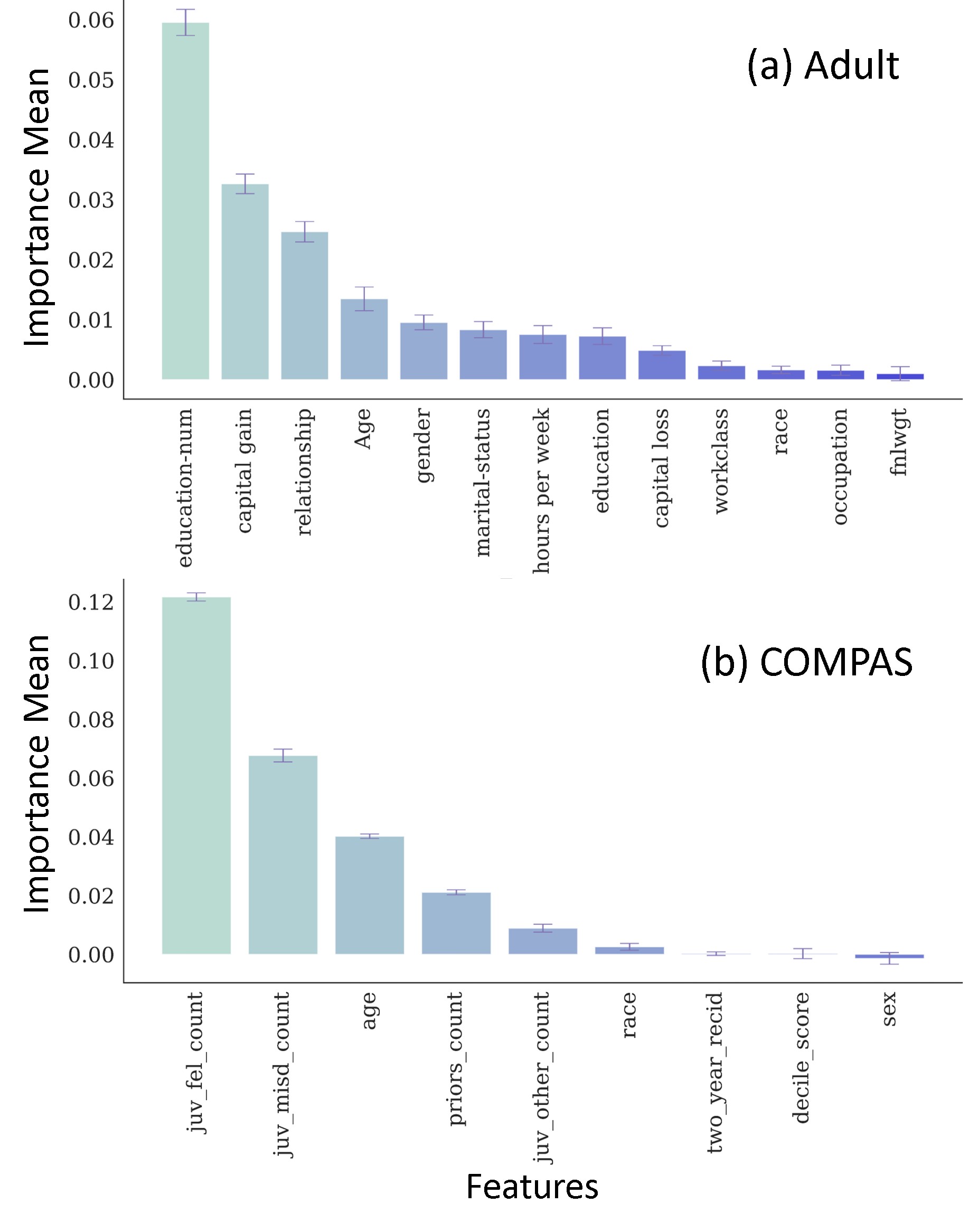}
    \caption{Critical features discovered by the \fairlearn.}
    \label{fig:crit_feats}
\end{figure}

One of the goals of the \fairlearn problem was to provide an interpretable way to estimate the features that contribute most to the overall outcome. To the best of our knowledge, we believe that our algorithm marks the first step in the direction of interpretable fairness in machine learning.We have previously argued that all features are not critically important for the final prediction made 
by the model. By extension, it also follows that some features which were thought to be the sensitive parameter did not exert a significant influence on the final outcome. Figure~\ref{fig:crit_feats} shows the critical features discovered by the \fairlearn algorithm for the Adult and the COMPAS dataset. The bars show the mean of five independent runs along with their standard deviations of the feature permutation algorithm. In the case of Adult dataset, we can see that the sensitive feature ``gender" is not as significant as other metrics like education and age. Interestingly, another potential sensitive feature ``race" has a very low importance. The critical features discovered for the COMPAS dataset are surprising. We find that the model places the \emph{least} importance on the ``Race" feature. On the other hand, the gender feature contributes more towards the overall outcome. Critical features of German dataset are not shown since we binarized the features into 59 categorical features. The most important features were found to be -  Property, Status, Credit History, Employment and Savings. The sensitive feature ``Sex" had an average feature importance of only $0.001$.

\section{Discussion}
\label{sec:discussion}
We discuss the main results of our work and possible future directions that emerge from it. 
\noindent

\textbf{Configurable Fairness}. The results of the previous section show that having the right setting can lead to significantly better fairness results. A key feature of our algorithm is to integrate flexible constraints into the optimization procedure. From our experiments, we recommend a range of $f_{tol} \in \left[10^{-2}, 10^{-3}\right]$ as a ``good" initial setting for optimization. In this paper we have shown an instance of optimization procedure using SVM, but in practice any classifier can be used once the appropriate objective has been defined. 

In our experiments we have shown a way to formulate a fairness metric into an optimization constraint and integrated it into a quadratic program to find decision boundaries using SVM. However, our algorithm is not limited by the choice of the machine learning algorithm. Once a fairness constraint has been translated into an optimization metric, it can be integrated with non-convex algorithms like neural networks using the method proposed in~\citep{Yang2020ProxSGD}. We leave this direction for future work. \\

\noindent



\textbf{Interpretability and Fairness}. The analysis of critical features in the previous section showed a surprising result - the attribute that was assumed to be the source of unfairness in the model, actually did not contribute anything to the overall prediction of the classifier. These incorrect assumptions about the sensitive attribute can be detrimental to the performance of the algorithm and can have unintended consequences in the real world. To the best of our knowledge, our method is the first to provide a principled approach for determining the sensitive parameter in the given dataset. We note that model-free methods of feature importance~\citep{fisher2019models} may yield better results.\\
 
\noindent

\textbf{Automatic Threshold Detection}. In certain scenarios there may not be a good setting for the tolerance parameter. In this case automated methods like~\citep{pmlr-v37-maclaurin15, pmlr-v70-franceschi17a} can be used to perform an iterative optimization of the fairness hyperparameter before the actual model training is started.

One potential strategy for automatic threshold detection can be \emph{hyperparameter optimization}~\citep{pmlr-v37-maclaurin15, pmlr-v70-franceschi17a, Bengio2000GradientBasedOO}. Given an initial estimate of $f_{tol}$ we can iteratively adjust it's value in the direction that leads to a lower amount of disparity between sensitive groups. In the case where the objective is convex (e.g. an SVM) we can solve a max-max optimization problem where we aim to maximize the dual of an SVM subject to a constraint that maximizes the distance of the sensitive attribute cluster from the decision boundary. In general, this question is an open problem and a rich area of research.


\section{Related Work}
\label{sec:related_work}
\noindent
In this section we discuss some related work in the development of fair classification algorithms.\\ 
\textbf{Fairness Definitions}. There have been a number of fairness definitions proposed for algorithmic fairness that address both group and individual fairness. The definitions of group fairness vary in the way the predictive measures for a classifier are considered. Almost all the definitions in group fairness are proposed for a binary classifier and a binary sensitive attribute. Two definitions of group fairness consider group parity to be tantamount to fairness. Disparate Impact~\citep{feldman2015certifying, zafar_fairness_2017} principle defines a classifier to be fair if the ratio of the positive prediction rates across sensitive groups is less than a certain threshold. Demographic Parity deems a classifier fair if the positive prediction rates are equal across sensitive groups. Both these metrics are defined for ideal cases and are hard to apply in the real world since datasets are rarely class balanced. Other metrics define fairness by considering the confusion matrix of a binary classifier and requiring that some performance metrics remain the same across sensitive groups. For instance, Equalized odds metric~\citep{hardt_equality_2016} requires that the true positive and false rates of a classifier are same across the sensitive group. Equal opportunity~\citep{hardt_equality_2016, donini_empirical_2020} relaxes this constraint by only requiring the true positive rates remain the same. A different definition of (un)fairness is provided by Zafar~\emph{et al}~\citep{zafar2017fairness} based on the misclassification rates in a binary classifier. A more thorough review of fairness metrics and their applications can be found in~\citep{Mitchell_2021}. One persistent issue in the algorithmic fairness community is that the fairness metrics while being internally consistent are inconsistent with other fairness metrics. Pleiss~\emph{et al.}~\citep{pleiss_fairness_2017} show that it is impossible to satisfy equalized odds at the same time except when the true positive and false negative rates are at zero. Dwork~\emph{et al.}~\citep{dwork2018group} also note the incompatibility of the fairness metrics under composition. Our work does not propose a new fairness metric, but it provides a way to integrate different metrics in a model agnostic fashion. This flexibility also makes our method future proof, since new metrics can be easily integrated into our algorithm.\\


\noindent
\textbf{Fair Optimization Strategies}. Fair classification algorithms can be grouped into three major categories depending on the stage in which they integrate the fairness constraint. The first category consists of algorithms that preprocess the data before running classification. Some examples of works that fall in this category are by Dwork~\emph{et al.}~\citep{dwork2018group} which proposes to learn a new data representation that removes the effect of protected attribute on the final decision. Kamiran~\emph{et al.}~\citep{KamiranFairness} also preprocess the data by changing the labels of highest ranking negative samples.
This ranking is learned by fitting a Naive Bayesian model on the dataset.
Friedler~\emph{et al.}~\citep{friedler2016impossibility} also define a data transformation method based on the Earth mover distance to mitigate disparate impact in algorithms. A drawback of methods in this category is that they assume that the classifier is inherently fair.  This prevents any measurement of the bias introduced by the model in the optimization process. Thus, even if the data is bias free, the resulting classifier may still output biased predictions due to it's internal bias. 
The second category of algorithms aims to integrate the fairness constraint during the training process of the classifier. Zafar~\emph{et al.}~\citep{zafar_fairness_2017} integrate the disparate treatment metric as a fairness constraint of a convex optimization problem. Donini~\emph{et al.}~\citep{donini_empirical_2020} present a way to integrate the equal opportunity metric as a linear constraint in the SVM objective. Kamishsima~\emph{et al.}~\citep{kamishima2011fairness} integrate fairness by adding a regularizer to the training objective that measures the degree of bias introduced by the algorithm. Dwork~\emph{et al.}~\citep{dwork_decoupled_2017} also propose a method which aims to train independent classifiers and then jointly fine-tune them to mitigate unfairness in the models. While this method does not integrate the fairness constraint into the training procedure, we put this method in this category since it mitigates fairness during an optimization procedure. The main advantage of these methods is that they allow for explicit modeling of fairness criteria and provide a way to control the amount of fairness in a model. Our work also falls in this category of algorithms since we expect to optimize for fairness during the training process. 
The third category of algorithms introduce fairness as a post-processing step. A  prominent work in this direction include the method by Hardt~\emph{et al.}~\citep{hardt_equality_2016} who introduce a post-processing step for the Equal Odds metric. This post processing takes the form of a linear optimization that is applied to the predictions made by a given unfair model. An advantage of methods that use a post processing step is that they are immensely flexible since they do not depend on the training data. Thus, they provide the most general methods of all the three categories and are useful in cases where the training data may not be available due to privacy concerns (e.g., medical record data). On the flip side however, there is no way to control the amount of fairness in the model which makes them useful in a limited number of applications as well as their start is always dictated by the initial solution of the original algorithm.

\section{Conclusion}
\label{sec:conclusion}

In this paper we introduced a training algorithm that provides a way to train fair classifiers in a configurable and interpretable manner. Our method is based on the observation that all features do not contribute equally to the final prediction and exploits a learned ranking of features to train models with top most predictive features. We decouple the fairness metric and the classifier thereby allowing a user to run optimization on the classifier with any fairness metric. Moreover, we allow the user to control the amount of fairness that can be introduced in the model through a tolerance parameter. As we observe in our experiments, our method consistently outputs better fairness results on different fairness metrics. We believe that our algorithm marks a step towards a unified view of fairness in machine learning.

Our work opens up many different directions of future research. One direction is to search for a more nuanced method of feature ranking. We would like to see the effect of Bayesian modeling in feature ranking and it's efficacy on fairness. A second direction is in the direction of optimal threshold detection for fairness. A hyperparameter optimization approach that detects the best tolerance parameter would be a welcome extension to our work. A relatively underexplored area in fairness literature is the optimization of algorithms with composite sensitive attributes. We define these attributes to be a combination of one or more sensitive attributes and jointly varying over their individual value. For example, if the sensitive attribute is race and gender, then a composite attribute is race-gender that takes all values in both race and gender combined. A future work direction then is to examine the critical feature hypothesis with composite attributes and extend our work to handle these kind of sensitive attributes. We note that multi-level SVM~\citep{sadrfaridpour2019engineering} is a way to improve the scaling of SVMs to large datasets. The adaptive SVM parameter training version \citep{aml-svm} takes the scalability one step further and could be an interesting approach to combine with the fariness parameter training. While our algorithm can be integrated directly into the framework, we leave the analysis of performance of our algorithm on this framework as future work.\\
\bibliography{fairness_ml}

\end{document}